\documentclass[letterpaper, 10 pt, conference]{ieeeconf}  

\IEEEoverridecommandlockouts                              

\usepackage{graphics} 
\usepackage{epsfig} 
\usepackage{mathptmx} 
\usepackage{times} 
\usepackage{amsmath} 
\usepackage{amssymb}  
\usepackage[ruled]{algorithm2e}

\title{\LARGE \bf
Reactive Multi-Robot Navigation in Outdoor Environments Through Uncertainty-Aware Active Learning of Human Preference Landscape}

\author{Chao Huang$^1$, Wenshuo Zang$^1$, Carlo Pinciroli$^2$, Zhi Jane Li$^2$, Taposh Banerjee$^3$, Lili Su$^4$, Rui Liu$^{1,*}$
\thanks{$^1$ are with the Cognitive Robotics and AI Lab (CRAI), College of Aeronautics and Engineering, Kent State University, Kent, OH 44240, USA. $^{*}$ Rui Liu is the corresponding author, email: ruiliu.robotics@gmail.com.
$^2$ are with Robotics Engineering, Worcester Polytechnic Institute, 100 Institute Rd, Worcester, MA 01609. 
$^3$ is with Industrial Engineering, University of Pittsburgh, 4200 Fifth Ave, Pittsburgh, PA 15260. 
$^4$ is with Electrical and Computer Engineering, Northeastern University, 360 Huntington Ave, Boston, MA 02115.}
}

\begin{document}

\maketitle
\thispagestyle{empty}
\pagestyle{empty}

\begin{abstract}
Compared with single robots, Multi-Robot Systems (MRS) can perform missions more efficiently due to the presence of multiple members with diverse capabilities. However, deploying an MRS in wide real-world environments is still challenging due to uncertain and various obstacles (e.g., building clusters and trees). With a limited understanding of environmental uncertainty on performance, an MRS cannot flexibly adjust its behaviors (e.g., teaming, load sharing, trajectory planning) to ensure both environment adaptation and task accomplishments. In this work, a novel joint preference landscape learning and behavior adjusting framework (\textit{\textbf{PLBA}}) is designed. \textit{PLBA} efficiently integrates real-time human guidance to MRS coordination and utilizes Sparse Variational Gaussian Processes with Varying Output Noise to quickly assess human preferences by leveraging spatial correlations between environment characteristics. An optimization-based behavior-adjusting method then safely adapts MRS behaviors to environments. To validate \textit{PLBA}'s effectiveness in MRS behavior adaption, a flood disaster search and rescue task was designed. 20 human users provided $1764$ feedback based on human preferences obtained from MRS behaviors related to \textit{\{"task quality", "task progress", "robot safety"\}}. The prediction accuracy and adaptation speed results show the effectiveness of \textit{PLBA} in preference learning and MRS behavior adaption.

\end{abstract}

\section{Introduction}
MRS has been widely used in complex real-world applications, such as disaster search-and-rescue \cite{9597860} and planetary exploration \cite{9438530}. However, deploying MRS in wide and unstructured environments remains challenging \cite{9447255}. A significant contributor is the difficulty in smoothly and seamlessly integrating human guidance into the robot's decision-making. But wide unstructured environments have various physical and logistical constraints \cite{9551457}, such as uncertain surroundings, limited communication, and unexpected obstacles. It's challenging for humans to gain precise understandings of priority and safety factors in missions and adjust the behavior of the MRS in real-time \cite{9310283}.

\begin{figure}[t!]
  \centering
  \includegraphics [width=1.0\columnwidth ]{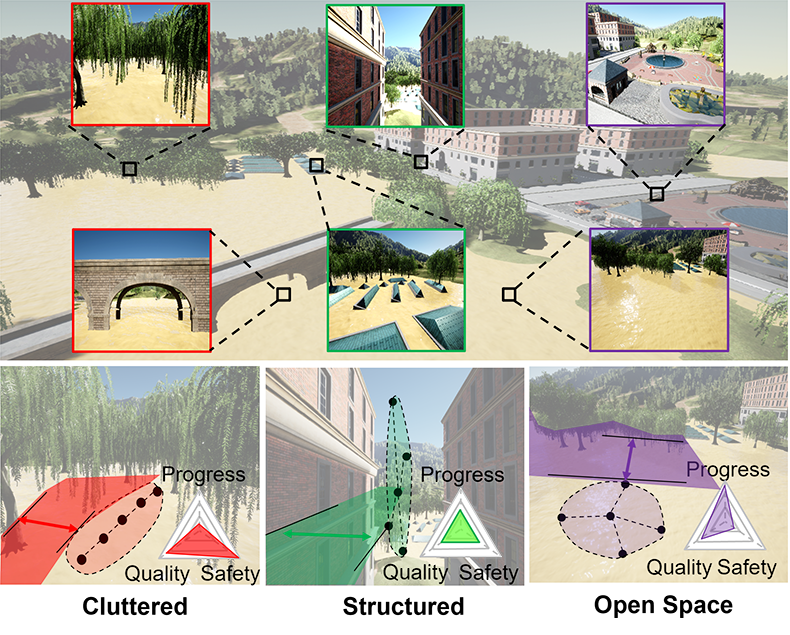}
  \caption{Illustration of a simulated flood disaster site. When deployed in a flood disaster response scenario, an MRS adapts to various environments, such as cluttered, structured, and open space environments, whose \textbf{characteristics} cannot be fully determined before actual deployment.}
  \label{idea}
\end{figure}

Currently, to enable more seamless integration and further improve robot survivability in real-world environments, many studies leverage machine learning methods to accurately predict human preferences for safe MRS deployments\cite{9020956}. Although the learned human preference model can provide robots with a fast and comprehensive understanding of human expectations, its learning and deployment still have the following flaws. 
First, human preference is dynamically associated with robot behaviors, task progress, and environmental conditions. Estimating human preferences by using only pre-defined motions and ignoring the above factors is challenging. For example, as illustrated in Fig. \ref{idea}, MRS deployed in open spaces flies fast and focuses more on task execution speeds, which is unacceptable for those deployed in a cluttered environment where robot safety is the primary consideration \cite{canal2019adapting}.
Second, humans may not always be capable of accurately determining their preferences, which results in uncertainty and noise in their feedback, hampering robots to utilize human preference for planning \cite{imtiaz2018sentiment}.

To improve the human-robot teaming with seamless human analysis and robot planning integrations, we proposed a novel joint preference landscape learning and MRS behavior-adjusting framework (\textit{PLBA}), shown in Fig. \ref{method}. In \textit{PLBA}, the Gaussian Process was used to quickly assess human preferences from noisy human feedback by leveraging spatial correlations between environment characteristics, and an optimization-based behavior control method was also used to ensure safe MRS behavior adjustments. As shown in Fig. \ref{idea}, \textit{PLBA} helps an MRS adapts to \{"cluttered", "structured", "open space"\} environments by balancing the task progress, execution quality, and robot safety. This paper has three major contributions:

\begin{enumerate}
\item A novel joint preference landscape learning and MRS behavior-adjusting framework (\textit{PLBA}) is developed to help robots understand relations between human preference landscape and environment's characteristics. This framework largely reduces robots' dependency on humans and improves robots' autonomy level.

\item An active collaboration-calibration scheme is designed to improve the human-MRS teaming. Instead of waiting for human corrections, \textit{PLBA} actively requests human guidance based on its learning needs (predicted uncertainty level), which can reduce the negative impact of feedback unreliability.


\end{enumerate}

\section{Related Work}

\textit{Robots Adaptation to Unstructured Environments}. 
In \cite{asadi2019real, trautmann2011mobility, 4651026}, a global environment classification model was learned to enable a robot to understand its surroundings and react in real-time.
In \cite{asadi2019real}, to develop a model for real-time perception and understanding of various environments, a lightweight neural network architecture was developed for real-time segmentation of navigable spaces.
To efficiently learn an environment characterization model, \cite{4651026} used the Probabilistic Gaussian method with sparse approximation to update parameters quickly. 
Although a global model can consider various features that may impact robots' movements, one of the main challenges in deploying global models is that global models are large and require a lot of computational and memory resources. As a result, global models can be slow to generate predictions, limiting their practical applications in real-time scenarios. 
Instead of learning a global model, \cite{levine2014learning, 6697168} developed several approaches to train an accurate local model for robot autonomous navigation. Even with local performance, these models need to be updated and retrained for each new environment. The model does not retrain well in new environments, which may lead to unsafe robot behavior. In environments where the trained model has low accuracy, our \textit{PLBA} method will seek guidance directly from the human operator, exploit the spatial correlations among similar environments for fast model updates, optimize robot behaviors, and improve planning safety even if the preference predictions are inaccurate.

\textit{Human Preference Model for Effective Humans-Robots Collaboration}. Recent research was done to learn human preferences to better integrate humans into robot teaming. In \cite{biyik2019asking, settles2009active}, active learning, addressing data inefficiencies by choosing the most informative queries, was used for robot learning as much as possible from a few questions. In \cite{biyik2018batch, 9081648}, the volume removal method, which aims to maximize the information gained from each query, was used to evaluate the performance of query methods. However, the above studies represented human preferences via a linear weighted sum of robot behaviors (e.g., moving speed, trajectory smoothness, and distance to obstacles), which can lead to the loss of valuable and informative information due to ignoring the dynamic and nonlinear influence entanglements among these behaviors \cite{9515509}. Furthermore, human preference is subjective and dynamically related to task progress and environmental conditions, making linear assumptions unsuitable for human preference representation \cite{9341530}. To solve the above issues, \cite{menard2021fast, krening2018interaction, 9369902} adopted methods based on reinforcement learning and neural networks to model human preferences. Even though neural networks can learn the complex, non-linear relationship between human preferences and robot behaviors, their performance was significantly impacted by the inherent noise present in human feedback \cite{imtiaz2018sentiment}. In response to this, our prior work\cite{huang2021meta} developed the Meta Preference Learning (MPL) method, which uses a meta-learning framework to adapt quickly to user preferences with minimal user input. This approach streamlines the integration of human guidance into MRS behavior adjustment, significantly reducing the cognitive load and frequency of human interventions required. In this work, to deal with varying levels of noise present in human feedback, the Gaussian Processes with Varying Output Noise is used to estimate the noise in the human feedback based on input. Then, the optimization-based MRS behavior adjustment approach can reliably and safely realize the predicted human preferences even with high-level uncertainty.

\section{Methods}

\subsection{Problem statement}
Human preference $H$ is parameterized by a set of actions $A $ with $ |A| = d $. Each action $a \in A$ represents one factor characterizing the multi-robot flocking process. In this work, multi-output Gaussian Processes (GPs) are used to model the underlying human preference function $f(\cdot) := [f^{(1)}(\cdot), f^{(2)}(\cdot), ..., f^{(d)}(\cdot)]^T$, which doesn't have strong assumptions about the form of $f(\cdot)$. Human preference feedbacks in trial $i$ are collected into a data set $Data_i = \{f_i^t + \epsilon_i^t | t = 1, 2, ..., N_p^i\}$, where $N_p^i$ denotes the total number of collected human feedbacks, $\epsilon_i^t \in \mathbb{R}^d$ represents noises in human feedback. Defining $\hat{f}(\cdot)$ as the maximum posterior (MAP) estimation of $f(\cdot)$ given $Data_i$, we aim at using a Bayesian approach to iteratively update the parameters of $f(\cdot)$ to minimize the prediction error ($|f(\cdot)-\hat{f}(\cdot)|$).

\subsection{Multi-output GPs with varying output noise}

 \begin{figure}[t!]
  \centering
  \includegraphics [width=1.0\columnwidth ]{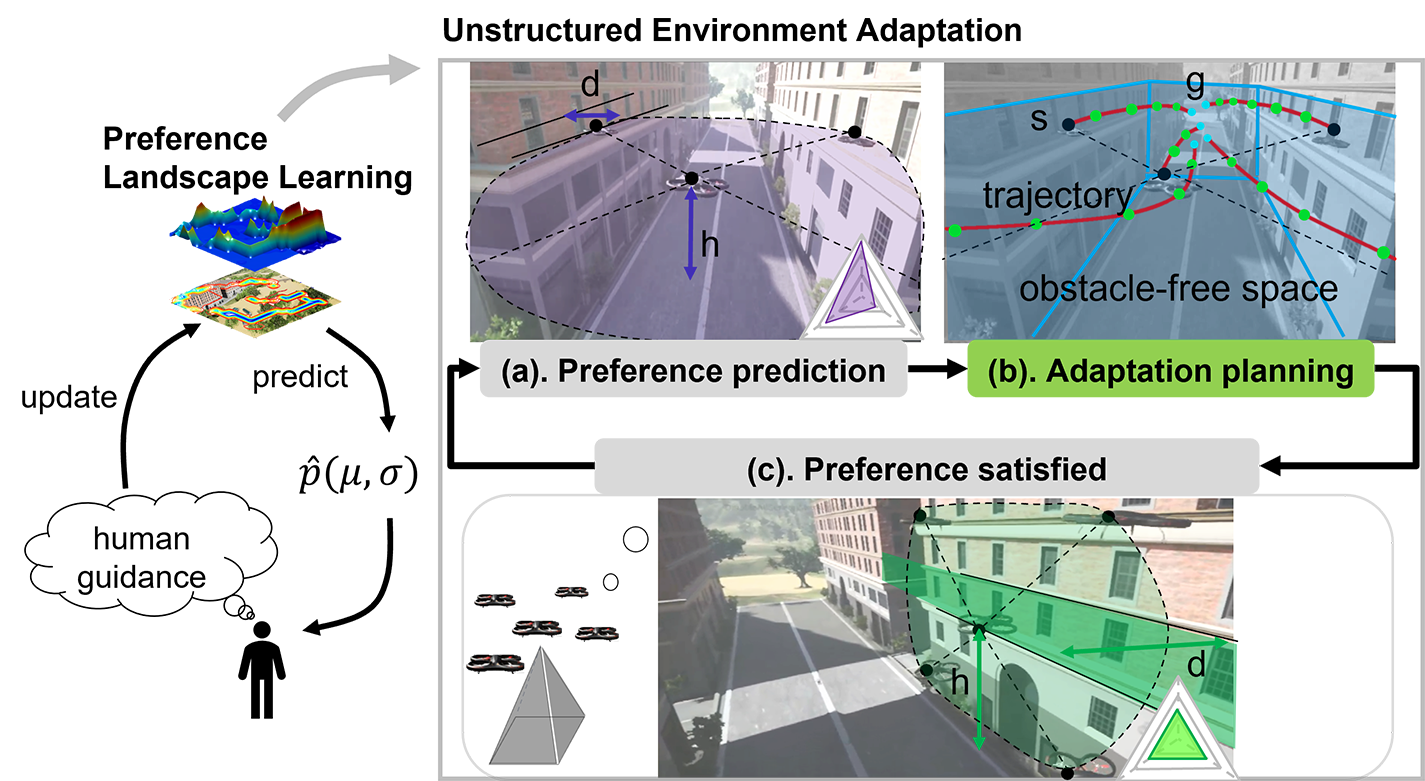}
  \caption{\textit{PLBA} workflow. The left part is the active learning framework based on preference learning. The right part is the unstructured environment adaptation process during MRS deployments. (a) preference prediction for environments, (b) human preference-based MRS behavior adaptation planning, (c) MRS behavior adjustment and preference model update.} 
  \label{method}
\end{figure}

After the problem statement, we are now ready to use multi-output Gaussian Processes \cite{van2020framework} to learn human preference function $f(\cdot): \mathbf{X} \to \mathbb{R}^d$. The $p^{th}$ output of $f(\cdot)$ is denoted as $f^{(p)}(\cdot)$. And the distribution of the stacked outputs ($F(X) = \{ f(x_n) \}_{n=1}^N$) for all $N$ inputs $x_n$ is Gaussian distributed, where $X=\{ x_n \}_{n=1}^N$ and the input $x_n \in \mathbf{X} \in \mathbb{R}^D$ is $D$-dimension combined features. Like single-output GPs, multi-output GPs are fully specified by their mean and kernel functions. The matrix-valued kernel obeys the same positive definiteness properties where the input
space $\mathbf{X}$ is extended by the index of the output: \begin{equation} k: (\mathbf{X},\mathbb{N}) \times (\mathbf{X},\mathbb{N}) \to \mathbb{R} \end{equation} Then the distribution of $F(X)$ can be denoted as: \begin{equation} p(F(X)) = \mathcal{N}(0, \mathbf{K}), \mathbf{K} \in \mathbb{R}^{(Nd \times Nd)} \end{equation} \begin{equation} F(X)_{np}=f^{(p)}(x_n), \mathbf{K}_{npn'p'}=k(\{x_n,p\},\{x_{n'},p'\}) \end{equation} Then, a linear transformation $W \in \mathbb{R}^{d \times L}$ of $L$ independent functions $g_l(\cdot)$ is used to construct multi-output function $f(\cdot)$: \begin{multline} g_l(\cdot) \sim GP(0,k_l(\cdot,\cdot')),\\ \mathbf{g}(x)=\{g_l(x)\}_{l=1}^L, f(x)=W\mathbf{g}(x)\end{multline} \begin{equation} k(\{x_n,p\},\{x_{n'},p'\})=\sum_{l=1}^LW_{pl}k_l(x_n,x_n')W_{p'l} \end{equation} where $k_l(\cdot,\cdot')$ is a squared exponential kernel.

Given that the human feedback for each output has a different noise level, a Gaussian process model where different noises are assumed for the data points is used: \begin{equation} y_n=f(x_n)+\epsilon_n, n=1,2,...,N \end{equation} \begin{equation} \epsilon_n|x_n \sim Gaussian(0,\Sigma_{noise,n}) \end{equation} In this paper, the covariance of noise ($\Sigma_{noise,n} \in \mathbb{R}^{d \times d}$) is represented as a diagonal matrix: \begin{equation}
    \Sigma_{noise,n}^{i,j} = 
    \begin{cases} e^{\alpha_{ij,n} + \beta_{ij,n}^Tx_n} & if(i==j) \\ 0 & otherwise \end{cases}
\end{equation} where $\alpha_{ij}$ and $\beta_{ij}$ are hyper-parameters related to human feedback noise. Then the distribution of $F(X)$ combined with noise is shown as follows:
\begin{equation}
    p(F(X))=\mathcal{N}(0, \mathbf{K}+\mathbf{K}_{noise})
\end{equation} where $\mathbf{K}_{noise} = diag(\Sigma_{noise,n})_{n=1}^N \in \mathbb{R}^{Nd \times Nd}$ is a diagonal matrix. With $d$-dimension outputs and $N$ samples, a $dN \times dN$ covariance matrix with $\Theta(d^3N^3)$ computational complexity is needed for obtaining an accurate multi-output GP. Hence, a posterior approximation method (stochastic variational inference) is used. By introducing a set of inducing variables ($\mathbf{u}=\{f(x_m)\}_{m=1}^M$), the computational complexity can be reduced to $\Theta(d^3M^3)$. At last, the Adam optimizer can optimize the noise variance and the variational parameters separately. Capturing the essence of our theoretical groundwork, Procedure 1 streamlines the execution of the PLBA framework (shown in Fig.  \ref{method}), from mission initiation to MRS safety assurance during ongoing operations. The algorithmic procedure outlined offers a succinct roadmap for dynamic decision-making, integrating path planning and real-time adaptive behavior modeling, which is pivotal for the MRS's performance in diverse environments. As for the optimization details, please refer to \cite{van2020framework, hensman2013gaussian}. 

\subsection{Preference-based safe MRS behavior adjustment}
In this work, we extended the flocking method proposed in \cite{vasarhelyi2018optimized} by involving obstacle-free largest convex region searching and optimization-based MRS behavior planning for stable and safe robot behaviors. The desired MRS behavior adjusting velocity is calculated by:
\begin{equation}
    v_i = v_i^{flock} + v_i^{fmt} + v_i^{rep} + v_i^{att} + v_i^{saf} + v_i^{hei} + v_i^{ali}
\end{equation}
where $i$ is the index of a drone and $v_i^{flock}$ denotes the flocking speed. $v_i^{fmt}$ represents MRS team formation maintaining term. $v_i^{rep}$ is short-range repulsion term and similarly $v_i^{att}$ denotes long-range attraction term. $v_i^{saf}$ keeps a safe distance between robots and obstacles. $v_i^{hei}$ maintains the robot at a preferred flying height. $v_i^{ali}$ is velocity alignment term. Then to prevent $v_i$ from largely exceeding the human preferred flying speed, an upper limit $h_{speed}$ is introduced:
\begin{equation}
    v_i = \begin{cases}\frac{v_i}{\| v_i \|}h_{speed} & if(v_i > h_{speed}) \\ v_i & otherwise \end{cases}
\end{equation}
Specifically, each type of speed can be parameterized by one or a few human preferences in the set $H = \{h_{inner}, h_{height}, h_{speed}, h_{safety}, h_{formation}\}$. $h_{inner}$ denotes the minimal distance between robots; $h_{height}$ denotes the flying height of robot system; $h_{speed}$ represents the flying speed of multi-robot system; $h_{safety}$ represents the minimal distance between robots and obstacles; and $h_{formation}$ denotes coordination pattern. Next, we determine the largest convex region of obstacle-free space using an iterative regional inflation method based on semi-definite programming \cite{Deits2015}. Drones are then assigned to the vertices of this convex region to establish the formation. The flocking model helps the drones maintain their relative positions within the formation throughout the operation.

\begin{algorithm}
\RestyleAlgo{ruled}
\caption{Pseudo Code of \textit{PLBA}'s Procedure}
 \While{Mission not finished \textbf{and} MRS is safe}{
  Obtain map data\;
  Path planning based on $A^*$ algorithm\cite{hart1968formal}\;
  \For{point in obtained path}{
  Predict $H$ and covariance from GPs\;
  \If{covariance $>$ threshold1}{
  Update $H$ to $H'$ with human guidance\;}
  Update local map\;
  Get the largest obstacle-free space\;
  result, loss $\gets$ MRS behavior adjust\; 
  \If{loss $\neq$ None}{
  \For{each MRS formation prototype}{
  result', loss' $\gets$ MRS behavior adjust\;
  \If{loss' $\neq$ None \textbf{and} loss' $<$ loss}{result $=$ result';
  loss $=$ loss'\;}}
  }
  Update $H'$ to $H''$ based on result\;
  MRS flocking control based on $H''$\;
  }
 }
 \label{alg:plba_algo}

\end{algorithm}

To maintain informative and helpful interactions with humans and learn a more representative preference model $f(\cdot)$ via human feedback (developed in Section A), Algorithm 1 was developed to decide the timing and manner of seeking human guidance. In Algorithm \ref{alg:plba_algo}, an interactive learning framework was designed by utilizing active learning to inquire informative guidance from humans. The mean and covariance of human preference predictions are calculated during the learning process. A high covariance value means that the human preference model does not generalize well to the current environment; thus, more human guidance is needed to refine the preference model. With the guidance of low-variance human preference, which means good preference estimation, MRS can adjust robot behavior safely without aggressive or unexpected adjustments.

\section{Evaluation}
A multi-robot search and rescue task in a flood disaster site was designed to evaluate \textit{PLBA}'s effectiveness in supporting unstructured environment adaptation. The following two aspects were mainly validated: (i) the effectiveness of \textit{PLBA} in fast human preference learning, which is for MRS fast adapt to the unstructured environment; (ii) the effectiveness of environment similarity for fast human preference learning. 

\begin{figure}[t!]
  \centering
  \includegraphics [width=1.0\columnwidth ]{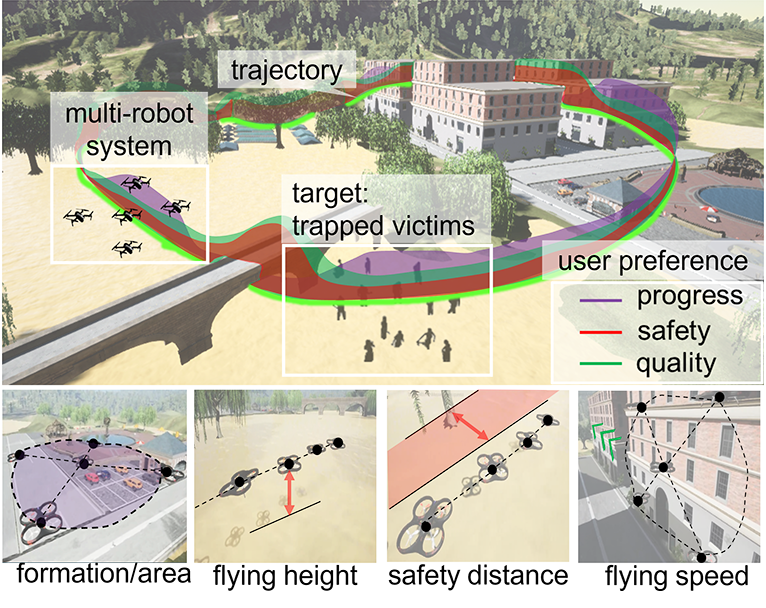}
  \caption{Illustration of experiment setting, including a flood disaster site, human preference (the width of trajectory is proportional to the ratio of three aspects of human preference), and multi-robot flocking behaviors.}
  \label{env}
\end{figure}

\subsection{Experiment setting}
In Fig. \ref{env}, an "MRS search and rescue in flood disaster" environment was designed based on the software AirSim \cite{shah2018airsim} and Unreal Engine \cite{games2019unreal}. The environment (400 $m$ \begin{math} \times \end{math} 400 $m$) was designed to include various scenes (city, bridge, forest, park, etc.), which is enough to test \textit{PLBA}'s effectiveness in supporting unstructured environment adaptation. Five kinds of flocking behaviors that can illustrate human preferences were considered: 1). The "flying height" denotes the average flying altitude of the MRS; 2). "Coverage area" is represented by the minimum distance between drones; 3). The "safe distance to the obstacle" is the minimum distance between the drone and obstacles; 4). "Flying speed" represents the average speed of MRS; 5). "Team formation" is represented by the assigned Cartesian positions of all robots.

In this work, outdoor environments mainly consist of four typical scenarios -- \textit{"City", "Park", "Forest", "County", and "River"}. According to obstacle density, these scenarios can be roughly classified into three categories \textit{"Cluttered", "Structured", and "Open Space"}. \textit{"Open Space"} represents a scenario where there are no or few physical obstacles (e.g., lake and park). In \textit{"Open Space"} environments, humans can operate MRS more aggressively to achieve fast mission progress.  \textit{"Cluttered"} represents areas that are characterized by an excessive accumulation of objects or debris, which can make it challenging to move around or perform activities (e.g., forest). Therefore, humans should operate MRS more reserved to pay more attention to robot safety. \textit{"Structured"} refers to spaces that are designed and organized with well-defined boundaries and guidelines, which are helpful to promote safety and consistency outcomes (e.g., city). Fig. \ref{beha} shows expected robot behavior corresponding to three specific environment types. 


To analyze the effectiveness of \textit{PLBA} in adaptation to outdoor environments, a test procedure for MRS to continuously adapt to multiple environments was designed; two baseline methods (linear regression and neural network regression) were used as comparison groups. The adaptation speed and prediction errors of human preferences in unfamiliar environments were comparatively analyzed. In addition, the relationship between environmental similarity and human preference prediction error was also analyzed, illustrating that the spatial correlation between environmental features can be used for environmental adaptation.

\subsection{Result analysis}

\begin{figure}[t!]
  \centering
  \includegraphics [width=1.0\columnwidth ]{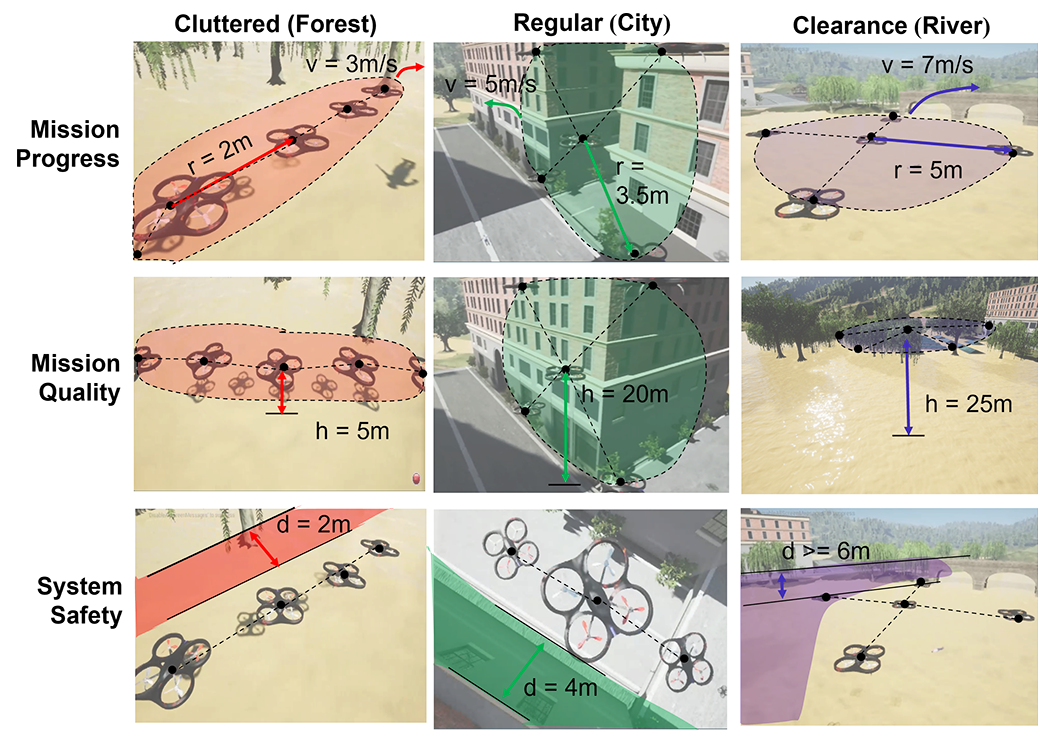}
  \caption{Preferred robot behavior illustration for three typical environments \textit{"Cluttered", "Regular" and "Clearance"}. }
  \label{beha}
\end{figure}

\begin{figure}[t!]
  \centering
  \includegraphics [width=1.0\columnwidth ]{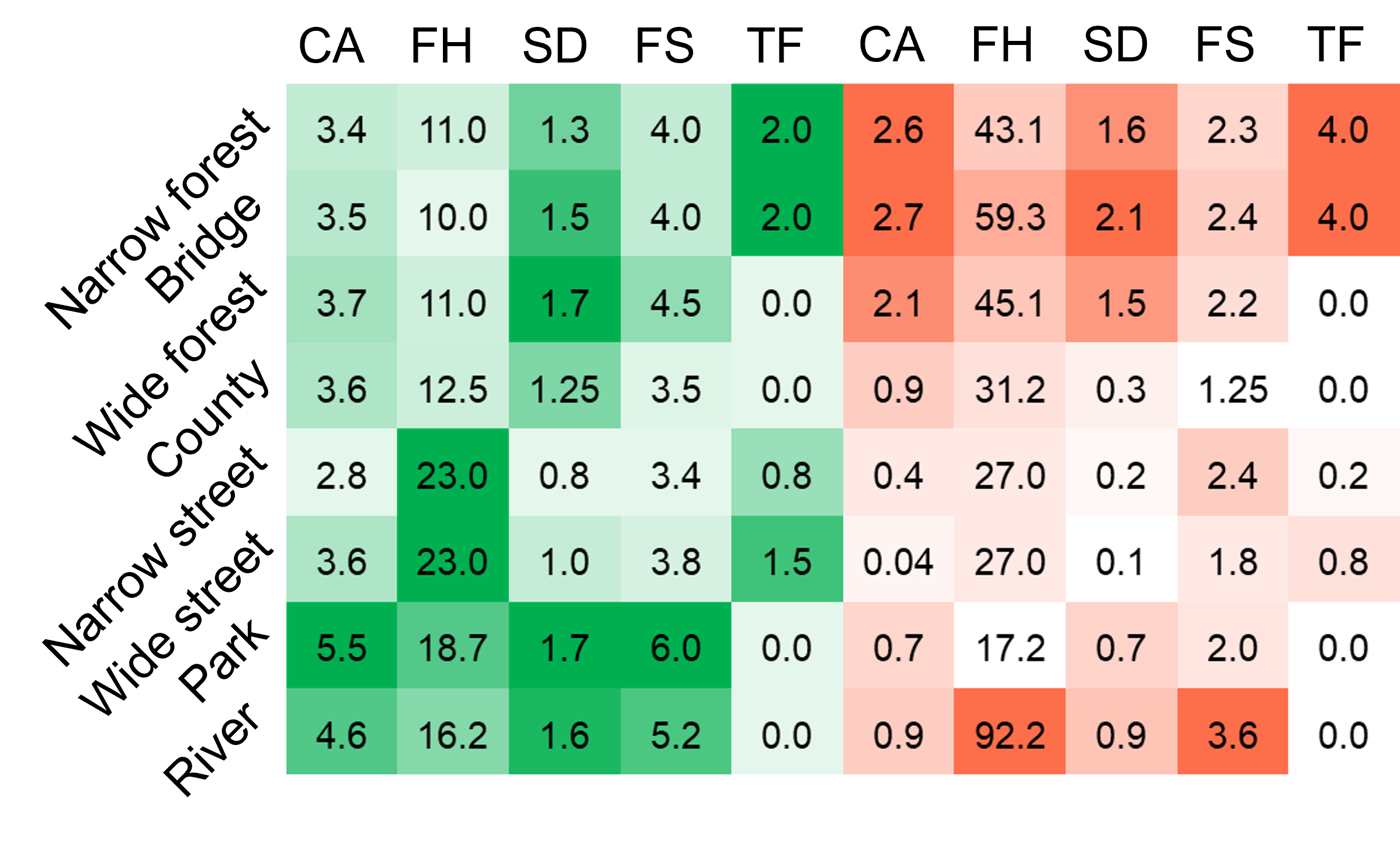}
  \caption{Human preference values for eight typical environments over five kinds of robot flocking behaviors ("CA": coverage area/m, "FH": flying height/m, "SD": safe distance to the obstacle/m, "FS": flying speed/(m/s), and "TF": team formation). Green represents the mean values of human preference, and red denotes the variance of human preferences.}
  \label{table}
\end{figure}

\textbf{Human User Study.} With the pioneer user study involving 20 human volunteers, the mean and variance human preference values in various environments were calculated. Then, two kinds of human preferences -- simple human preference only with mean values and complex human preference with mean and variance values were conducted to validate the effectiveness of \textit{PLBA} in rapid adaptation to unstructured environments. The details of human preferences are listed in Fig. \ref{table}. The sample behavior illustrations for human preferences in different environments are shown in Fig. \ref{beha}. 

\begin{figure}[t!]
  \centering
  \includegraphics [width=1.0\columnwidth ]{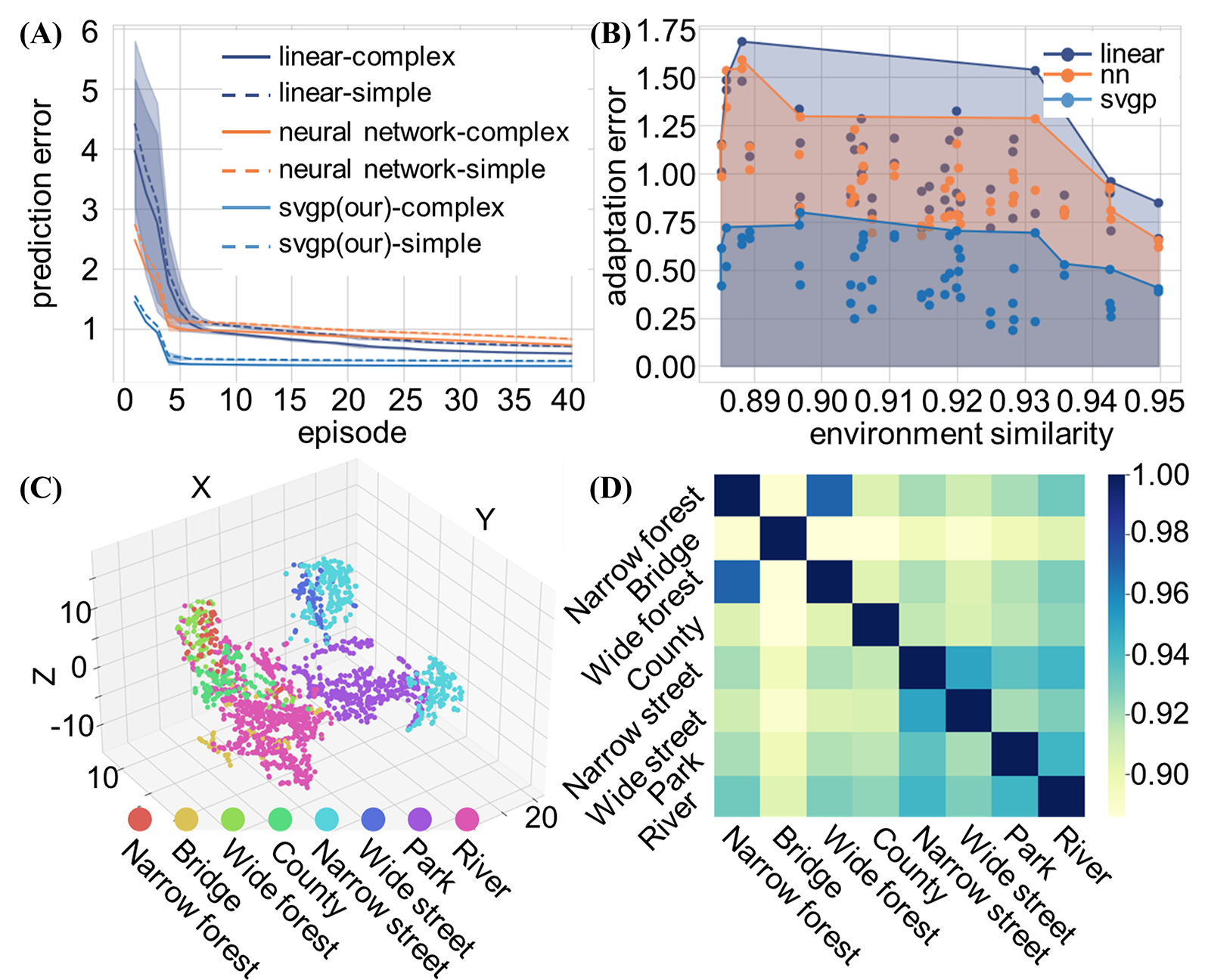}
  \caption{(A) Testing Loss comparison between the developed method \textit{PLBA} and the baseline methods -- linear regression and neural network regression. It shows that with the same update number, \textit{PLBA} adapted to human preference faster than baseline methods. (B) Furthermore, the spatial correlation between environmental features benefits fast unstructured environmental adaptation. (C) Environment feature clustering based on t-SNE. (D) The similarity between different environments.}
  \label{loss}
\end{figure}

\textbf{Environment analysis and learning Performance Validation.} 
Under the guidance of volunteers, human preference over motion behaviors such as \{"covering area", "flying height", "distance to an obstacle", "flying speed", and "team formation" \} were collected to generate a data set with $1764$ samples; the data set is mainly sampled from a flood disaster site which can be mainly categorized into eight typical environments. Each category includes about 12.5\% of the total data samples. Each sample consists of environment features and corresponding human preference guidance. The environment features are concatenated by environment visual features, MRS status, and the relationship between MRS and the environment. The distribution and similarity of various environments are illustrated in Fig. \ref{loss} (C) and (D). As shown in Fig. \ref{loss} (A), after five updates, the $L2$ losses between predicted human preference and real human preference generated by \textit{PLBA} is less than 0.05; while the $L2$ losses obtained from baseline methods are greater than 1.00. THey show the proposed method \textit{PLBA} has a faster convergence speed to customize MRS motions for unstructured environment adaptation.

\begin{figure}[t!]
  \centering
  \includegraphics [width=1.0\columnwidth ]{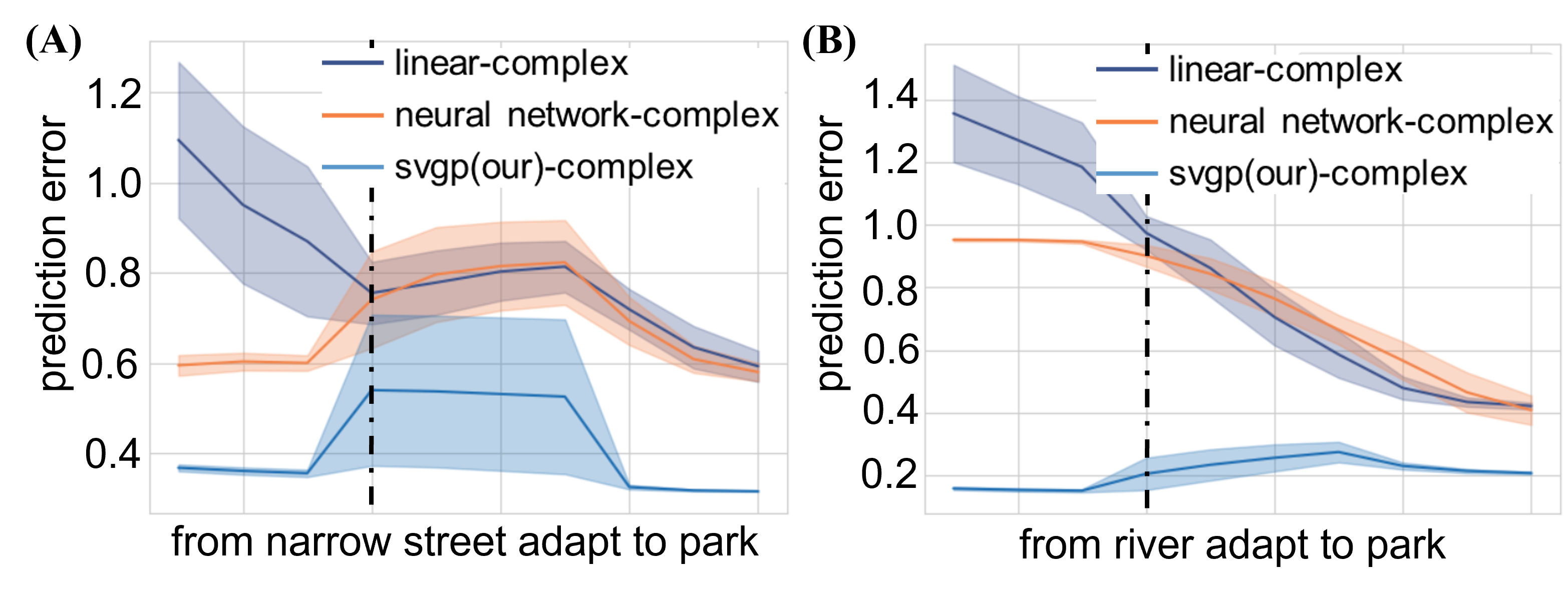}
  \caption{(A) Adaptation between dissimilar environments. (B) Adaptation between similar environments.}
  \label{process}
\end{figure}

\textbf{Adaptation Effectiveness Validation.} The relationship between the prediction errors and environment similarity was analyzed to validate the method's effectiveness in facilitating adaptation of unstructured environments. The rationale is that when two environments are similar, the prediction errors of a model trained in an old environment but used in new environments will be minimal.
Two examples of changes in human preference prediction errors from one environment to another new environment were illustrated. 
As shown in Fig. \ref{loss} (B), the human preference adaptation error will decrease with the increase of similarity between the previous and new environments for our proposed method and all the baseline methods. That means the spatial correlation between environments can be leveraged for fast unstructured environment adaptation. Besides, the adaptation error of \textit{PLBA} is 0.5 less than that of baseline methods, which means our proposed method is more suitable for fast and accurate unstructured environment adaptation.
Fig. \ref{process} shows two samples of new environment adaptation. (A) The human preference prediction error for MRS to adapt from a narrow street to a park environment is around 0.5; (B) while the human preference prediction error for MRS to adapt from a river to a park environment is around 0.3. That means the model trained in a river environment can quickly adapt to the park environment because the part environment is more similar to the river environment than the narrow street environment, verified in Fig. \ref{loss} (D).

\section{Conclusion and Future Work}
This paper developed a novel joint preference landscape learning and MRS behavior-adjusting framework (\textit{PLBA}) to establish a mapping between human preference landscape and unstructured environment's characteristics and adjust MRS motions to follow human preferences over mission progress, quality, and system safety. To validate method effectiveness, a five-UAV-based multi-robot team was deployed for victim search in a flood disaster site; with 20 volunteer-based user studies, two types of user preferences were identified as "simple-preference" -- without variances and "complex-preference" -- with variance. Our proposed method's effectiveness for unstructured environment adaptation was validated by the reduced update number and the adaptation error calculated by models updated only once in a new environment. Given the capability of enabling safe and cognitive teaming, \textit{PLBA} can be extended to guide the flexible teaming of multiple robots and even the cooperation between vehicles and human units. 

In the future, the adaptation method could be enriched by on-board sensing information to improve human-MRS collaboration further.



\end{document}